# Analysis of the Reasoning with Redundant Information Provided Ability of Large Language Models

Wenbei Xie

Beijing-Dublin International College, Beijing University of Technology, 100124, China

wenbei.xie@ucdconnect.ie

**Abstract.** Recent advancements in Large Language Models (LLMs) have demonstrated impressive capabilities across a range of natural language processing tasks, especially in reasoning, a cornerstone for achieving Artificial General Intelligence (AGI). However, commonly used benchmarks may not fully encapsulate the inferential abilities of these models in real-world scenarios. To address this gap, a new form of Question-Answering (QA) task, termed Reasoning with Redundant Information Provided (RRIP), is introduced. The study designed a modified version of the grade school math 8K (GSM-8K) dataset which has several variants focusing on different attributes of redundant information. This investigation evaluates two popular LLMs, LlaMA2-13B-chat and generative pre-trained transformer 3.5 (GPT-3.5), contrasting their performance on traditional QA tasks against the RRIP tasks. Findings indicate that while these models achieved moderate success on standard QA benchmarks, their performance notably declines when assessed on RRIP tasks. The study not only highlights the limitations of current LLMs in handling redundant information but also suggests that future training of these models should focus on incorporating redundant information into the training data to increase the performance on RRIP tasks.

## 1. Introduction

Recent large language models (LLMs) have made large improvements across a wide range of Natural Language Processing tasks [1][2]. Among these improvements, the improvements in reasoning tasks are impressive since inferential ability is one of the most important abilities to achieve Artificial General Intelligence (AGI). Among recent LLMs that made improvements on reasoning tasks, one of the most shining stars is generative pre-trained transform 4 (GPT-4), which achieves State-Of-The-Art on several mainstream academic benchmarks [3]. Investigating the benchmarks listed in the GPT-4 technical report, it can be seen that most of the mainstream benchmarks chosen to investigate inferential ability focus on two types of Question-Answering (QA) tasks: One is that all of the information given in the question is needed to give the correct answer to the question, like GSM-8K and MATH [4][5]. Another is that language models need to choose the correct answer between multiple choices by only using their pre-trained knowledge, like High School European History task in MMLU dataset [6].

However, it can be one-sided evaluation criteria to evaluate the inferential ability of a language model by only using these two types of QA tasks. Generally, these forms are derived from the forms of exam questions for human students. For the purpose of giving a clear indication to the students, the first question form basically does not contain redundant information in questions. The second question form generally aims to investigate whether students understand a specific concept. To solve this kind of question, complex reasoning is basically not required. These forms of questions can be suitable for examination, but most of the challenging real-world problems are not shown in these forms. These challenging real-world problems can be considered as having two noticing features: First, complex reasoning ability is needed to solve the problems. Second, large amounts of information can be related to the question, while only limited pieces of information are actually needed to solve the question. This form of QA task can be described as Reasoning with Redundant Information Provided (RRIP). If one LLM can achieve high accuracy on RRIP, it can be thought a step closer to achieving AGI since most of the innovations are generated by solving the challenging real-world problems mentioned above.

In fact, the questions in some of the popular datasets like SQuAD and Natural Questions have presented the existence of redundant information [7][8]. For instance, Natural questions dataset ask language models to find answers of the given question in a given Wikipedia page. To solve questions in this form, language models have to find useful information in the Wikipedia page and utilise it to reason about the true answer. However, complex reasoning ability is not the focus of this type of datasets, and the influence different attributes may appear on the redundant information are also not addressed.

The primary goal of this study is to evaluate the ability of RRIP of two LLMs based on QA tasks, LlaMA2-13B-chat and generative pre-trained transformer 3.5 (GPT-3.5) [9]. LlaMA2-13B-chat is a popular lightweight pre-trained model which is widely used in industry, while GPT-3.5 is the base model of ChatGPT, which is also widely used by individuals and businesses. By investigating the RRIP ability of these two popular language models, a brief overview of the RRIP ability of current LLMs can be obtained. It is shown that LlaMA2-70B got 56.8% accuracy and GPT-3.5 got 57.1% accuracy on the GSM-8K dataset [3][4][9]. By using this moderate accuracy as a baseline, it is relatively easy to overserve visible change of accuracy after having modified questions in grade school math 8K (GSM-8K) into questions with redundant information. To clearly demonstrate the effects made by different types of redundant information, several variants of information have been designed to check the influence of different features of redundant information. The experiment result shows that the accuracy of RRIP is generally lower than the accuracy on baseline. Based on the answers given by LLMs, this study also evaluates the appearance and reasons for performance loss to provide information on the future improvement of the RRIP ability of LLMs.

## 2. Methodology

*2.1. Dataset description and preprocessing*
The Grade School Math 8K (GSM-8K) dataset is a math problem dataset released by OpenAI. It contains 8.5k well-designed primary school math problems written by human writers [4]. For each problem, a detailed answer which contains several steps of inference is provided. The dataset also provides a second form, which is called Socratic Dataset. In this form, auto-generated sub-questions and answers are provided along with each question by decomposing them. In this research, only the basic form of GSM-8K has been used. To achieve the research purpose, manual modification has been applied to 50 question-answer pairs in the GSM-8K dataset to generate several variants of questions that have redundant information with different features.

*2.2. Proposed approach*
The major objective of this study is to give a comprehensive evaluation of the RRIP ability of two popular LLMs, LlaMA2-13B-chat and GPT-3.5. The general process to achieve the objective is described in Figure 1. First, manual modification is applied to the questions in the GSM-8K dataset. Then, models are asked about the questions to get the answer given. Finally, the results are analysed.

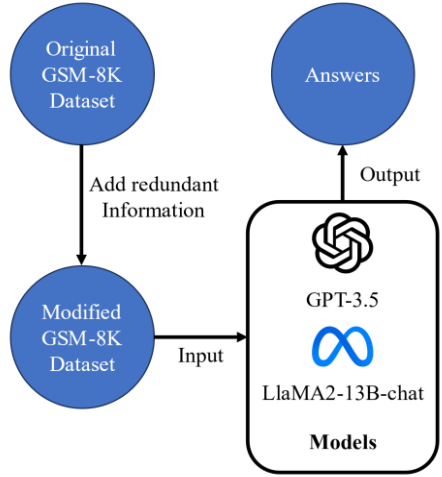

**Figure 1.** Process of the study

*2.2.1. Redundant information.* To this end, a clarification on the concept of redundant information is needed first. A typical question in the GSM-8K dataset and similar datasets consists of two parts, one is the sentences that provide information, which is described as information I in this paper, and another is the interrogative sentence that actually gives the question, which is described as the question Q in this paper. Figure 2A shows an example of this typical structure. Generally, the information provided in these datasets is just enough to help an agent to gain the answer to the question. Thus, the information can be described as sufficient condition to answer the question, denoted as information set S in Figure 2. Based on this, redundant information can be clearly defined. Redundant information means some of the information that cannot provide any help to achieve the answer. That means redundant information is neither sufficient nor necessary condition for the question, which is denoted as information set N in Figure 2B.

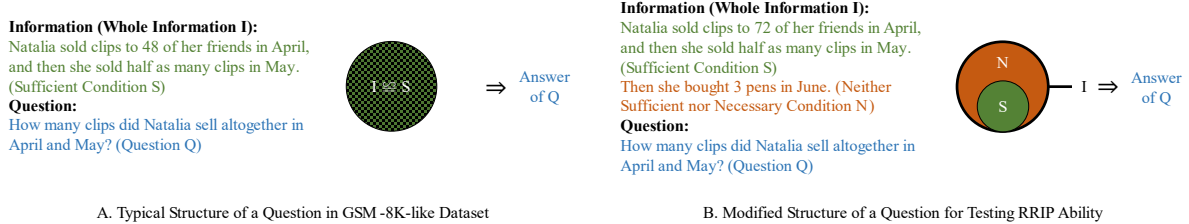

A. Typical Structure of a Question in GSM -8K-like Dataset    B. Modified Structure of a Question for Testing RRIP Ability

**Figure 2.** Comparison between typical math problem dataset and RRIP dataset

*2.2.2. Types of redundant information.* This study designed several different types of redundant information to figure out the effect of the following factors. To investigate the effect of relevance between N and S, four different types of redundant information have been designed as Figure 3: (1). Duplication (DU), where the redundant information set N was generated by simply duplicate information set S. (2). Simplification (SI), where the previous S information set was turned into I which contains information set S' and information set N by simplify the question Q. (3). Same Subject with Number (SSN), where the redundant information sets N was designed to have same subject as information set S. Furthermore, numbers would appear in the information set N. (4). Meaningless Information (MI), where the redundant information sets N was designed to have no relationship with information set S since the redundant information came from articles written by British Broadcasting Corporation (BBC) travel. Among the four types of redundant information, DU is the type that most related to the question in the original dataset, and MI is the type that is least related to the original question. However, these two types of redundant information are easier to be handled by language models. Although SSN and SI are more related to the question in the original dataset, both of them are kind of confusing, which means even humans have to read the questions carefully to pick out useful information. To investigate the effect of combining Named Entity Recognition (NER) task with math problem solving task, a special form of redundant information has been constructed. To construct this type of redundant information, the information set S was copied at first. Then, the subject in S' was changed into another subject which was randomly given. Finally, the numbers in S' were also changed to get the redundant information set N. It can be considered that the RRIP task has been transformed into the Reasoning while Recognising Named Entity (RRNE) task.

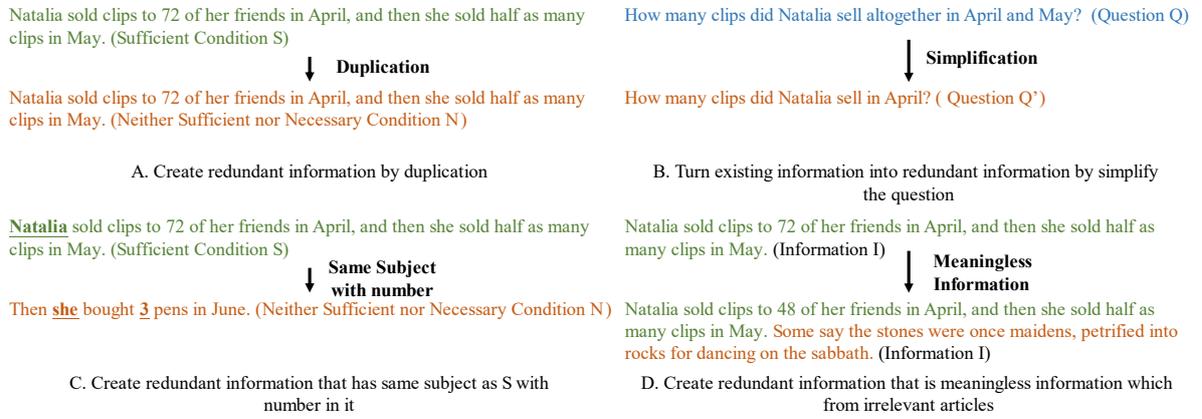

**Figure 3.** Example of four types of redundant information with different relevance

*2.2.3. Evaluation metrics.* Since there is uncertainty in the model output due to sampling methods chosen by the model creator, the evaluation of the output accuracy should be reconsidered. First, the judgment standard on whether an output is correct or not should be clarified. In this paper, if the output contains the right answer and the output does not rewrite the answer after giving the right answer, the model can be regarded as making a correct answer. Based on this, each question has been asked to the models for five times to check the consistency of the answer correctness. 3 evaluation metrics have been developed based on this, including accuracy (ACC), single-right rate (SRR) and almost-right rate (ARR).

Accuracy (ACC) represents the percentage of correct answers among all the answers given by the model, it can be calculated by the formula below:

$$Accuracy = \frac{Number\ of\ correct\ answers}{Total\ number\ of\ answers} \quad (1)$$

It provides a valuable assessment of how possible the model tends to give correct answers.

The single-right rate (SRR) evaluates whether a model has the least probability of giving correct answers. it can be calculated using the following formula:

$$SRR = \frac{q_1 + q_2 + q_3 + q_4 + q_5}{Total\ number\ of\ questions} \quad (2)$$

where $q_n$ $(0 \leq n \leq 5, n \in N)$ represents the number of questions that have n correct answer given by models.

The almost-right rate (ARR) evaluates whether a model has the ability to constantly give correct answers. It can be calculated using the formula below:

$$ARR = \frac{q_5}{Total\ number\ of\ questions} \quad (3)$$

where $q_n$ $(0 \leq n \leq 5, n \in N)$ represents the number of questions that have n correct answer given by models.

### 2.3. Implemented details

The study ran inference using LlaMA2-13B-chat-hf which comes from hugging face. To obtain different answers for one question among different turns of chat, LlmMA2's temperature was set to 0.8, top-k was set to 50 and top-p was set to 0.95. Furthermore, to keep the level of basic accuracy, the "Let's Think step by step" prompt has been always added to the system prompt for both models [10].

## 3. Result and discussion

### 3.1. Effect of relevance

Figure 4 presents the change of the three metrics after applying redundant information which has different relevance to the original information.

For LlaMA2-13B-chat as Figure 4A, all three metrics have suffered losses on DU, MI and SSN. Especially when LlaMA2-13B-chat gives answers to the question with SSN redundant information in it, it suffers a nearly 10% loss of accuracy. One possible reason could be the model tends to take the numbers in the redundant information into the calculation. These numbers even can cause hallucination which is presented as making up non-existent numbers to help the calculation. Although the accuracy of SI has slightly increased, it still can be considered a loss of accuracy. The reason will be discussed below. After simplifying the question, it is much easier to answer the question. Thus, it ought to have a large increase in accuracy instead of a slight increase.

For GPT-3.5 (Figure 4B), the losses on all three metrics are different. The drop in accuracy only appears on SSN and SI, which is more than 10%. The same reason as the LlaMA2's loss on the SSN task can be applied to explain the drop in GPT-3.5's performance on SSN and SI.

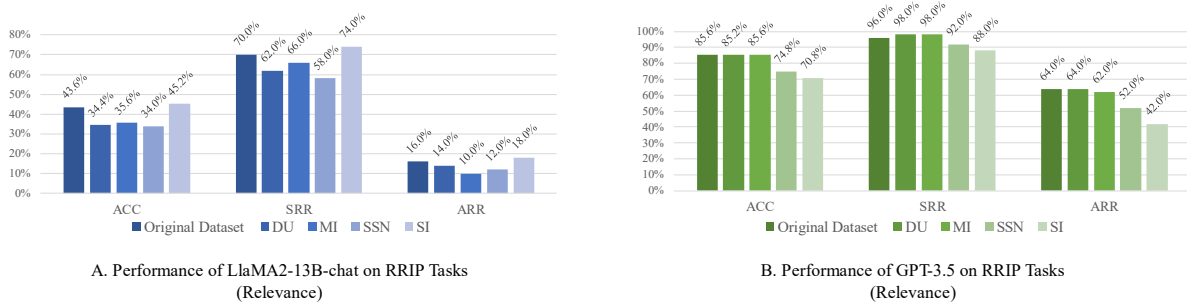

A. Performance of LlaMA2-13B-chat on RRIP Tasks (Relevance)

B. Performance of GPT-3.5 on RRIP Tasks (Relevance)

**Figure 4.** Performance of models on relevance-related task group

### 3.2. Change of performance when processing RRNE task

Figure 5 presents the difference in the three metrics between processing the RRNE task and processing original dataset.

For both LlaMA2-13B-chat (Figure 5A) and GPT-3.5 (Figure 5B), the values for RRNE are lower than the values for the original dataset on all three metrics. A reason similar to the reason for accuracy loss on SSN can cause such situation. When model deals with the redundant information, it tends to calculate based on it despite the instruction of question. This calculation may reduce the attention model paid on the needed information, which causes the loss of accuracy.

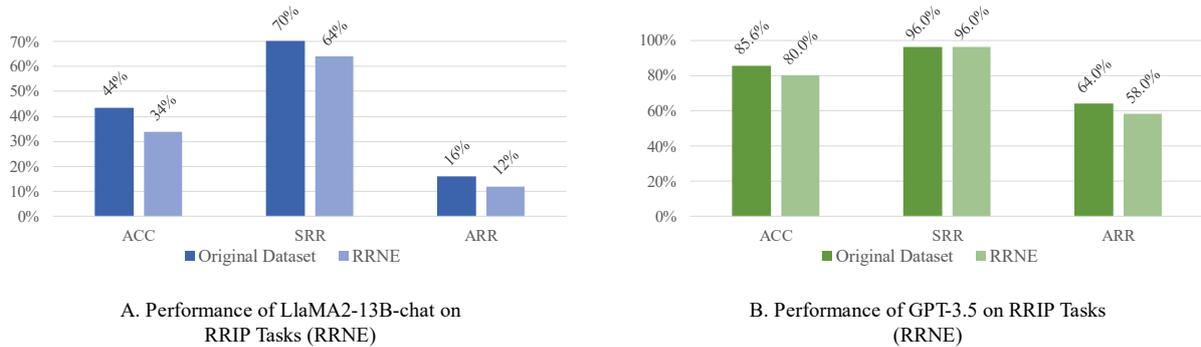

A. Performance of LlaMA2-13B-chat on RRIP Tasks (RRNE)

B. Performance of GPT-3.5 on RRIP Tasks (RRNE)

**Figure 5.** Performance of models on RRNE task

### 3.3. Overview of the change in performance

In general, both LlaMA2-13B-chat and GPT-3.5 have generally suffered a loss on all three metrics (Figure 6), although the overall values of GPT-3.5 are higher than LlaMA2-13B-chat on all three metrics. When dealing with redundant information which cannot cause much confusion, like DU and MI, GPT-3.5 appears to be more robust than LlaMA2-13B-chat. This may be caused by the large parameter amount of GPT-3.5. However, when processing SI, LlaMA2-13B-chat performs much better than GPT-3.5. Furthermore, the losses on SRR of LlaMA2-13B-chat are generally greater than GPT-3.5's, which may indicate that LlaMA2-13B-chat tends to lose the basic probability of giving correct answers. In contrast, the loss on ARR of LlaMA2-13B-chat on SSN and SI are generally smaller than GPT3.5's, which may show that Chat-GPT tends to be less confident to give correct answers constantly when dealing with these types of redundant information.

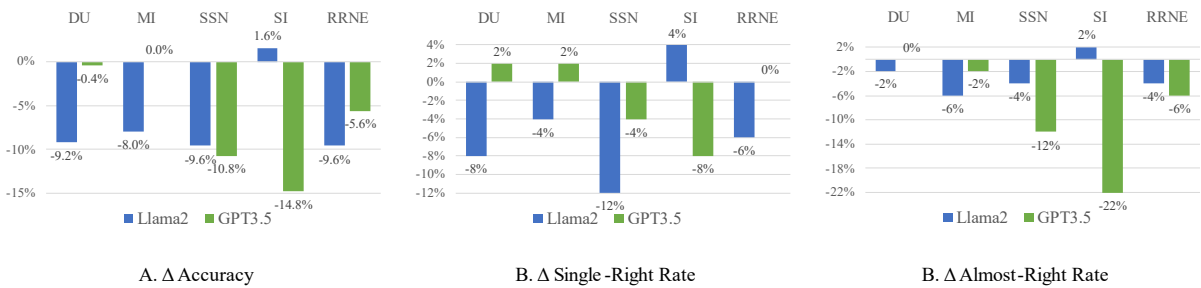

A. Δ Accuracy

B. Δ Single-Right Rate

B. Δ Almost-Right Rate

**Figure 6.** Delta of all three metrics

## 4. Conclusion

To examine the ability of popular LLMs to reason with redundant information provided, this study designed several types of redundant information and modified the GSM-8K dataset to construct a comprehensive evaluation scheme on the ability. The results show that the performance of LLMs is lost when processing questions which have redundant information, and the performance decreases more as the redundant information becomes more confusing, especially the redundant information that has the same subject as the original information with numbers in it. The direct reason for the performance loss is that LLMs tend to take all the numbers that appear in the question into consideration and calculation, which may allocate limited attention to unnecessary calculations. Another phenomenon shown by the result is that the performance of LLMs is also lost when dealing with the combined task. Based on the results, there is still a long way for LLMs to achieve the objectives to become AGI. For future research, a larger sample size and evaluation of the influence of the position of redundant information can be performed to better understand the effect of redundant information. To solve the problem mentioned in this study, future training on LLMs should focus on this problem by constructing part of the training data with redundant information provided.